\documentclass[final,3p,times]{article}

\usepackage[a4paper,left=25mm,top=25mm,right=20mm,bottom=35mm]{geometry}

\usepackage{authblk}
\usepackage[numbers, sort&compress]{natbib}
\usepackage{amsmath,amssymb} 
\usepackage{graphicx}
\usepackage{tabularx}
\usepackage[lofdepth,lotdepth]{subfig}
\usepackage{xcolor}
\usepackage{url}

\usepackage[colorlinks]{hyperref}
\hypersetup{
	citecolor=blue,
   linkcolor=red,
}
\usepackage[margin=30pt,font=small,labelfont=bf]{caption}

\providecommand{\keywords}[1]{\textbf{\textit{Keywords: }} #1}
   
\title{A Weighted Population Update Rule for PACO Applied to the Single Machine Total Weighted Tardiness Problem}

\author[1]{Daniel Abitz}

\author[2]{Tom Hartmann}

\author[1]{Martin Middendorf}

\affil[1]{Swarm Intelligence and Complex Systems Group, Faculty of
    Mathematics and Computer Science, University of Leipzig, Augustusplatz
    10, D-04109 Leipzig, Germany.}

\affil[2]{Bioinformatics Group, Department of Computer Science \&
    Interdisciplinary Center for Bioinformatics, Universit{\"a}t Leipzig,
    H{\"a}rtelstra{\ss}e~16--18, D-04107 Leipzig, Germany.}

\date{\ }

\begin{document}
\sloppy

\maketitle

\abstract{ 
In this paper a new population update rule for population based ant colony optimization (PACO) is proposed. PACO is a well known alternative to the standard ant colony optimization algorithm. The new update rule allows to weight different parts of the solutions. PACO with the new update rule is evaluated for the example of the single machine total weighted tardiness problem (SMTWTP). This is an $\mathcal{NP}$-hard optimization problem where the aim is to schedule jobs on a single machine such that their total weighted tardiness is minimized. PACO with the new population update rule is evaluated with several benchmark instances from the OR-Library. Moreover, the impact of the weights of the jobs on the solutions in the population and on the convergence of the algorithm are analyzed experimentally. The results show that PACO with the new update rule has on average better solution quality than PACO with the standard update rule.
}


%
%
	\section{Introduction}
	\label{sec:intro}
		Population based ant colony optimization (PACO) algorithm \cite{GM2002} is an iterative metaheuristic where a population of solutions is transferred from one iteration to the next iteration. In each iteration the  population is used to generate corresponding pheromone information which is then used by the ants as in standard ant colony optimization (ACO) in order to construct new solutions. An advantage of PACO is that it exhibits faster pheromone update and evaporation mechanisms than usual ACO algorithms while being competitive with respect to solution quality (e.g., \cite{GM2002,Oliveira2011,weise2014tspsuite}). For more information on different ACO approaches as well as recent
		developments in the research field of ACO metaheuristics the reader is referred to \cite{DS2004} and \cite{Dorigo2019}, respectively.
		
        Since the population in PACO determines the pheromone information the PACO metaheuristic uses a population update rule to change the pheromone values. Thus, the population update rule in PACO corresponds to the pheromones update rule of ACO. In this paper a new population update rule for PACO is proposed that uses weights for different parts of the solutions in order to control the strengths of their influence on the optimization process. PACO with the new update rule (WPACO) is applied to the single-machine total weighted tardiness problem (SMTWTP). The SMTWTP is a well-studied scheduling problem that is known to be $\mathcal{NP}$-hard \cite{LKB1977}. An instance of the SMTWTP is a set of jobs where each job has a processing time, a due date, and a weight. The aim is to find a schedule of the jobs on a single machine such that the weighted total tardiness, i.e., the weighted sum of delays caused by finishing a job after its due date, is minimized.
        
        The principle of the new update rule (when applied to the SMTWTP) is to incorporate the weights of the jobs into the PACO algorithm. Instead of a population of solutions, the WPACO uses a sequence $(P_1,\ldots ,P_n)$ of multisets of jobs, where $n$ is the number of jobs. Multiset $P_i$ contains the jobs that were on place $i$ in the best solutions that were found by the ants in the last iterations. 
        In PACO it is a principle that the iteration best solution is entered into the population.
        In WPACO a corresponding principle is used: For each position $i\in [1,n]$ of the iteration best solution, the job $j$ on position $i$ enters the multiset $P_i$.
        Each multiset $P_i$ has a maximum capacity $k$ and the sum of the weights of the jobs that are stored in the multiset cannot exceed $k$. Details of the new update rules are described in Section~\ref{sec:aco}.
	
        Algorithm WPACO is evaluated on several benchmark instances from the OR-Library \cite{ORLIB} and is experimentally compared to PACO with the standard population update rule. For the experiments the parameter values of WPACO are optimized using the automatic configuration tool \texttt{Irace} \cite{IRACE}.
		
		The paper is organized as follows: in Section~\ref{sec:smtwtp} a formal definition of the SMTWTP is presented.
		In Section~\ref{sec:aco} ACO and PACO are described as well as the new weighted population update rule.
		The experimental setup is given in Section~\ref{sec:exp} and discussed in Section~\ref{sec:results}. In Section~\ref{sec:conc} a summary of the paper is given and avenues for future work are outlined.
		
%
%
	\section{Single Machine Total Weighted Tardiness Problem}
	\label{sec:smtwtp}
		The \emph{single machine total weighted tardiness problem (SMTWTP)} is defined as follows:
		Consider a set of $n\in\mathbb{N}$ jobs that need to be processed on a single machine that can handle at most one job at a time. Each job $j$ is
		assigned to a \emph{processing time} $p_{j} \in \mathbb{N}_{\geq 0}$ that describes the time that is needed to process job $j$, a \emph{due date}
		$d_j\in \mathbb{N}_{\geq 0}$ that describes the time point when the processing of job $j$ should have been finished, and a \emph{weight} 
		$w_j\in \mathbb{N}_{\geq 0}$ that represents the priority of job $j$. Given such a set of $n$ jobs, a \emph{schedule} $\pi$ is a permutation of
		length $n$, i.e., a bijective mapping  $\pi:\! \{1,...,n\} \rightarrow \{1,...,n\}$,  that assigns to each place $i$ in the queue a job $\pi(i)$.
		For the sake of a clear notation, we represent a permutation $\pi$ as the $n$-tuple $(\pi(1),...,\pi(n))$.
		Clearly, a schedule $\pi$ defines a total order in which the $n$ jobs are processed on a single machine with $\pi(1)$ (respectively $\pi(n)$)
		being the first (respectively last) job in $\pi$. 
		For a given schedule $\pi$, the \emph{completion time} $C_{j}$ of a job $j$ is the time that is needed to complete job $j$ 
		in $\pi$, i.e., $C_j:=\sum_{i=1}^{i \leq \pi^{-1}(j)}p_{\pi(i)}$, where $\pi^{-1}(j)$ denotes the position of job $j$ in $\pi$.
		The \emph{tardiness} $T_j$ of a job $j$ is defined as $T_j:=\max\{C_j - d_j,0\}$. 
		Note that the tardiness cannot be negative and thus, it can be seen as a penalty for completing a job after its due date.
		Given a set of $n$ jobs, the single machine total weighted tardiness problem aims to find a schedule of all $n$ jobs that minimizes the 
		weighted tardiness of all jobs, i.e., it aims minimize the objective $\sum_{j=1}^n w_jT_j$.
		The expression $\sum_{j=1}^n w_jT_j$ of a schedule $\pi$ is also called the \emph{total weighted tardiness} of $\pi$.  
		
		If $w_1=...=w_n=1$, then the objective function of the SMTWTP can be simplified to $\sum_{j=1}^n T_{j}$.
		This problem is called single machine total tardiness problem.
%
%
	\section{Ant Colony Optimization for the SMTWTP}
	\label{sec:aco}
		This section presents background information on different ACO approaches for the SMTWTP. In particular,
			in Section~\ref{subsec:ACO} the ACO approaches proposed in \cite{BSD2000} and \cite{MM2000} are described.
			In Section~\ref{subsec:PACO} the PACO metaheuristic that has been presented in \cite{GM2002} is outlined. 
			The details of the proposed weighted population update rule for PACO are described in Section~\ref{subsec:WPACO}.
		
		\subsection{ACO for the SMTWTP}
		\label{subsec:ACO}

		In this section the ACO approach for the SMTWTP of Besten et al. \cite{BSD2000} is described.
		Consider a given SMTWTP with $n$ jobs to be scheduled. Recall that a schedule of the $n$ jobs is represented as a permutation $(\pi(1),...,\pi(n))$. 
		For example the permutation $(4, 1, 3, 2)$ is a schedule of $4$ jobs in which job $4$ is processed first, followed by jobs $1$, $3$, and $2$ in that order.
		The ACO approach of Besten et al. is initialized with a fixed number of ants and number of iterations.
		In each iteration every ant starts with an empty schedule and iteratively appends unscheduled jobs until the schedule is complete,
		i.e., all jobs are scheduled.
		Through this process it is ensured that all jobs are scheduled and no job is scheduled multiple times.
		Two kinds of information are used in order to influence an ant's decision to select a job \textit{j} at position \textit{i}.
		The heuristic information $\eta_{ij}$ indicates how desirable it is to schedule job $j$ at position $i$ with respect to a problem specific
		heuristic function. The pheromone value $\tau_{ij}$ gives details about favorable schedules that have been found in previous iterations.
		A large value $\tau_{ij}$ indicates that of job $j$ has often been placed at position $i$ by ants in previous iterations. This implies that placing
		job $j$ at position $i$ may be favorable with respect to the objective function, since only "good" schedules of former iterations are usually
		allowed to update the pheromone values.
		For the solution construction, Besten et al. combine a maximization strategy with a probabilistic decision. 
		Consider a probability parameter $q_0$ with $0 \leq q_{0} < 1$.
		With probability $q_0$ an ant chooses a job $j \in S^{*}$ from the set of unscheduled jobs $S^{*}$ at position $i$ if and only if the expression
		\begin{equation}
			(\tau_{ij})^{\alpha} \cdot (\eta_{ij})^{\beta}
			\label{eq:nextjob_max}
		\end{equation}
		is maximized, where $\alpha$ and $\beta$ are parameters that represent the influence of the pheromone information and the heuristic information,
		respectively.
		With probability $(1 - q_{0})$ job $j\in S^*$ is scheduled on position $i$ randomly with respect to the probability
		\begin{equation}
			p_{ij} = 
			\begin{cases}
				\frac{(\tau_{ij})^{\alpha} \cdot (\eta_{ij})^{\beta}}
						 {\sum_{k \in S^{*}} (\tau_{ik})^{\alpha} \cdot (\eta_{ik})^{\beta}} & \text{, if $j \in S^{*}$}\\
				0 & \text{, otherwise.}
			\end{cases}
			\label{eq:nextjob_prob}
		\end{equation}
		
		In \cite{MM2000} it has been shown that the ACO algorithm from Besten et al. can significantly be improved.
		The central idea of such an improvement is to used sums of pheromone values instead of a single pheromone value in order to guide an ant's decision.
		This mechanism allows the ants to consider pheromone values that have already been used for making earlier decisions.
		Clearly, the authors of \cite{MM2000} proposed the following summation strategy: 
		For position $i$ all preceding pheromone values should be considered and thus, they suggested using the sum of pheromone values of all already
		scheduled jobs. Using the ideas presented in \cite{MM2000} formulas~\eqref{eq:nextjob_max} and \eqref{eq:nextjob_prob} can be updated
		resulting into the following ACO algorithm.
		With probability $q_{0}$ an ant chooses a job $j \in S^{*}$ at position $i$ that maximizes
		\begin{equation}
			(\sum_{l = 1}^{i} \tau_{lj})^{\alpha} \cdot (\eta_{ij})^{\beta}
			\label{eq:nextjob_sum_max}
		\end{equation}
		and with probability $(1 - q_{0})$ job $j \in S^{*}$ is chosen at position $i$ randomly with respect to the probability
		\begin{equation}
			p_{ij} = 
			\begin{cases}
				\frac{(\sum_{l = 1}^{i} \tau_{lj})^{\alpha} \cdot (\eta_{ij})^{\beta}}
						 {\sum_{k \in S^{*}} (\sum_{l = 1}^{i} \tau_{lk})^{\alpha} \cdot (\eta_{ik})^{\beta}} & \text{, if $j \in S^{*}$}\\
				0 & \text{, otherwise.}
			\end{cases}
			\label{eq:nextjob_sum_prob}
		\end{equation}
		
		Different heuristics can be used in order to compute the heuristic information $\eta_{ij}$.
		The most basic one is called the Earliest Due Date (EDD). The EDD heuristic prefers jobs that have a small due date and thus,
		the heuristic values are calculated by $\eta_{ij}=1/d_{j}$.

		A more elaborated heuristic called the Modified Due Date (MDD) heuristic has been proposed in \cite{BBHS1999} and was 
		further improved in \cite{MM2000}.
		The MDD heuristic considers (in addition to the due date) the potential completion time $C_j$ a job $j$ would have if scheduled at position $i$.
		Following this notion, the heuristic values are calculated by
		\begin{equation}
			\eta_{ij} = \frac{1}{max\{C_j,d_{j}\} - (C_j - p_j)}.
			\label{eq:mod_mdd}
		\end{equation}
	
		After an ant has scheduled a job, the following local pheromone update is performed:
		The pheromone value $\tau_{ij}$ is replaced with $(1 - \rho)\tau_{ij} + \rho \tau_{0}$, where $0\leq \rho < 1$ is a given evaporation 
		parameter and $\tau_{0}$ an initial amount of pheromones. Value $\tau_0$ is computed with respect to the SMTWTP as $\tau_0=1/(nT_{EDD})$,
		where $n$ is the number of jobs and $T_{EDD}$ is the total tardiness of a schedule obtained via the EDD heuristic.
		Note that the value $\tau_0$ is also used to initialize the pheromone matrix.
		
		At the end of each iteration, i.e., after all ants have constructed a schedule for all jobs, the global amount of pheromone is updated by means
		of two procedures. First, pheromone is evaporated by setting $\tau_{ij}$ to $(1 - \rho) \tau_{ij}$.
		The idea behind evaporation is that the influence of old solutions is reduced during the run of the algorithm.
		The second procedure performs an additional update of the global pheromone values for all job-position pairs that occur in the best schedule that has been found so far. 
		In detail, if job $j$ is at position $i$ in the best schedule $\pi$,
		then the pheromone value $\tau_{ij}$ is increased by $1/T_{b}$, where $T_b$ is the weighted tardiness of $\pi$.
	
		The algorithm stops after a termination criterion is met, e.g., a specific number of iterations is reached.
	\subsection{PACO for the SMTWTP}
		\label{subsec:PACO}
		In this section, the population based ant colony optimization approach (PACO) for the SMTWTP that has been presented 
		in \cite{GM2002} is described. Generally, the PACO algorithm follows the same procedure as the algorithms explained in
		Section~\ref{subsec:ACO}: In each iteration a fixed number of artificial ants construct a new solution, pheromone is
		updated, and pheromone is transmitted to the following iteration. 
		However, the pheromone update and the transmission are different.
		Instead of using a matrix of pheromone values that is transmitted from one iteration to another, PACO uses a set of solutions,
		called a population, from which the pheromone values can be calculated in each iteration. In addition, pheromone update in the PACO algorithm
		is performed by changing the solutions in the population. Compared to the ACO, the pheromone update of the population based approach is much
		faster \cite{Oliveira2011}. The following paragraph describes these procedures of the PACO algorithms in detail,
		see also Figure~\ref{fig:paco} for an example.
	
		Consider a population $P$ with a capacity of $k\in\mathbb{N}$ schedules, i.e., $P=\{\pi_1,...,\pi_h\}$ with $0\leq h\leq k$.
		The pheromone values $\tau_{ij}$ that are needed for constructing a new solution, i.e., formulas~\eqref{eq:nextjob_sum_max} and
		\eqref{eq:nextjob_sum_prob}, can be computed by $\tau_{ij}=\tau_0 + \tau_s\ell_{ij}$, where  $\tau_{s} = (\tau_{max} - \tau_{0})/k$ and $\ell_{ij}\in\{1,...,h\}$ denotes how often job $j$ is at position $i$ in the $h$ schedules contained in $P$. 
		Parameter $\tau_{max}$ controls the maximal amount of pheromones.
		Formally, the value $\ell_{ij}$ is defined as $\ell_{ij}:=|\{\pi \in P: \pi(i)=j\}|$, where $|X|$ denotes the cardinality of a set $X$.
		Note that this implies that for each pheromone value $\tau_{ij}$ it holds that $\tau_{0} \leq \tau_{ij} \leq \tau_{0} + k\tau_s$,
		and thus $\tau_{0} \leq \tau_{ij} \leq \tau_{max}$.
		At the end of each iteration, i.e., after all artificial ants have constructed a new schedule, evaporation is performed by removing the
		oldest schedule from the current population. This population update rule is called age-based strategy \cite{GM2002} and it is the
		standard rule for PACO. In addition, the iteration-best schedule is added to the population. At the beginning of the
		optimization process, either the initial population is empty or it is filled with $k$ schedules that are constructed randomly or heuristically.
		If the initial population is empty, then no schedule is removed from the population in the first $k$ iterations.
		See Figure~\ref{fig:paco} for an example of the population update rule of the PACO algorithm.
		
		\begin{figure}
			\centering
			\subfloat[][]{
				\begin{tabular}{|cccc|}
					\hline
					(\textbf{3}, & \textbf{1}, & \textbf{2}, & \textbf{4}) \\\hline
					(3, & 2, & 1, & 4) \\\hline
					(4, & 2, & 1, & 3) \\\hline
					- & - & - & -
					\\\hline
			\end{tabular}}
			\qquad
			\subfloat[][]{
				\begin{tabular}{|cccc|}
					\hline
					(\textbf{1}, & \textbf{4}, & \textbf{2}, & \textbf{3}) \\\hline
					(3, & 1, & 2, & 4) \\\hline
					(3, & 2, & 1, & 4) \\\hline
					(4, & 2, & 1, & 3)
					\\\hline
			\end{tabular}}
			\qquad
			\subfloat[][]{
				\begin{tabular}{|cccc|}
					\hline
					(\textbf{2}, & \textbf{3}, & \textbf{4}, & \textbf{1}) \\\hline
					(1, & 4, & 2, & 3) \\\hline
					(3, & 1, & 2, & 4) \\\hline
					(3, & 2, & 1, & 4)
					\\\hline
			\end{tabular}}
			\caption{
				The figure shows the standard population update of the PACO algorithm. Subfigures~(a) to (c) illustrate
				population $P$ with a capacity of $4$ at the end of iteration~$3$ to $5$, respectively.
				Each line within $P$ is either empty (dashes) or it is filled with a schedule.
				The schedule that was added to $P$ during the respective iteration is highlighted.
				(b) Population $P$ after schedule $(1, 4, 2, 3)$ was added.
				(c) Since the capacity of $P$ is reached, the oldest schedule $(4, 2, 1, 3)$ is removed from $P$.
				In addition, the schedule $(2, 3, 4, 1)$ is added to $P$.
			}
			\label{fig:paco}
		\end{figure}
		\subsection{Weighted population update rule for PACO}
		\label{subsec:WPACO}
			A novel population update rule for the PACO algorithm for the SMTWTP is proposed in this section. 
			The idea of this rule is to consider the weights of the jobs of a schedule that is added to the population. 
			We refer to the PACO algorithm that uses the novel population update rule as weighted population based ant colony optimization (WPACO) algorithm.
			Generally, the WPACO algorithm follows the same procedures as the PACO algorithm that is described in Section~\ref{subsec:PACO} but it differs in the way a population is construed and the way a schedule is added to the population at the end of an iteration. 
			Instead of considering a population $P=\{\pi_1,...,\pi_k\}$ of capacity $k\in\mathbb{N}$, the idea of the novel population update rule in the WPACO
			algorithm is to consider a \emph{weighted population} $wP=(P_1,...,P_n)$ that contains for each position $i\in \{1,...,n\}$ a multiset
			of jobs $P_i=\{\pi_1(i),...,\pi_k(i)\}$ with capacity $k$ that were scheduled to position $i$ in the last $k$ iterations.
			Observe that this is a difference to the notion of the PACO algorithm, since it has the benefit that a population is no longer bound to contain feasible schedules. 
			Whereas this may appear counterproductive, it allows to perform an update rule that considers the weight of a job of an SMTWTP as explained in the following.
			
			Consider an SMTWTP of $n\in\mathbb{N}$ jobs and a weighted population $wP=(P_1,...,P_n)$ in which each multiset has a capacity of $k\in\mathbb{N}$. 
			Recall that each job $\pi(i)$ is assigned to a weight $w_{\pi(i)}$ that represents its priority.
			Suppose that schedule $\pi$ is added to $wP$ at the end of an iteration and job $\pi(i)$ is scheduled to position $i$ in $\pi$, then evaporation is performed by removing the oldest $w_{\pi(i)}$ jobs from multiset $P_i$. 
			In addition, $\pi(i)$ is
			added $w_{\pi(i)}$ times to $P_i$ in order to fill the weighted population.
			At the beginning of the optimization process, the WPACO behaves analogously to the PACO algorithm.
			In particular, each multiset of the weighted population is either empty or it is filled with jobs of schedules that were constructed randomly or heuristically. 
			Likewise, if the initial weighted population is empty, then no job is removed from a multiset $P_i$ until $P_i$ is completely filled. 
			Figure~\ref{fig:wpaco} illustrates an example of this novel population update rule.

			\begin{figure}
				\centering
				\subfloat[][]{
					\begin{tabular}{|c|c|c|c|}
						\hline
						3 & 2 & 1 & 4 \\\hline
						4 & 2 & 1 & 3 \\\hline
						- & - & - & - \\\hline
						- & - & - & -
						\\\hline
				\end{tabular}}
				\qquad
				\subfloat[][]{
					\begin{tabular}{|c|c|c|c|}
						\hline
						\textbf{3} & \textbf{1} & \textbf{2} & \textbf{4} \\\hline
						\textbf{3} & 2 & \textbf{2} & 4 \\\hline
						\textbf{3} & 2 & 1 & 3 \\\hline
						3 & - & 1 & -
						\\\hline
				\end{tabular}}
				\caption{Weighted Population $wP=(P_1,...,P_4)$ with a capacity of $4$ at the start~(a) and end~(b) of an iteration of the WPACO algorithm.
				The multisets $P_1,...,P_4$ of $wP$ are illustrated by the columns (from left to right), i.e., $wP=(P_1=\{3,4\},P_2=\{2,2\},P_3=\{1,1\},P_4=\{4,3\})$.
				The weights of the four jobs in the exemplified SMTWTP are $w_1=w_4=1$, $w_2=2$, and $w_3=3$.
				The schedule $(3, 1, 2, 4)$ is added to $wP$ at the end of this iteration.
				Since the capacity of $P_1$ is $4$, the oldest job $4$ is removed. 
				The reason is that adding job $3$ three times to $P_1$ would result	in a multiset that contains 5 jobs.
				Observe that no job was removed from $P_2$, $P_3$ and $P_4$ as they satisfy this capacity constraint.
				Subsequently, job $3$, $1$, $2$, and $4$ is added $3$, $1$, $2$, and $1$ times to $P_1$, $P_2$, $P_3$, and $P_4$, respectively.
				The added jobs are highlighted in~(b).
				}
				\label{fig:wpaco}
			\end{figure}			
					
			Figures~\ref{fig:paco} and \ref{fig:wpaco} demonstrate the primary difference between the population update rule in the PACO algorithm and the WPACO algorithm: Using the novel update rule and weighted populations, a population may contain partial
			and invalid solutions. For example, only the first row in Figure~\ref{fig:wpaco}.(b) represents a feasible schedule for the given SMTWTP.
			The reasoning is that jobs occur multiple times in the remaining rows. 
			However, it is worth to mention that this does not affect the way artificial ants construct their schedules. 
			The reasoning is that the artificial ants of the PACO and the WPACO algorithm construct their schedules with respect to the pheromone
			values $\tau_{ij}$ which are depend on the values of parameters $\tau_0$, $\tau_s$ and on the value $\ell_{ij}$ that represents how often
			job $j$ is at position $i$ in the current population. It is not hard to see that the value of $\tau_0$ and $\tau_s$ can be set easily.
			In addition, the value $\ell_{ij}$ can also be obtained from a weighted population $wP=(P_1,...,P_n)$ by counting how often
			job $j$ occurs in the multiset $P_i$. Consequently, the artificial ants of the WPACO algorithm construct feasible solutions as well as the ones of the PACO algorithm.

%
%
	\section{Experiments}
	\label{sec:exp}
		SMTWTP instances from the OR-Library \cite{ORLIB} were used to investigate the optimization behavior of the proposed weighted population update rule. 
		An SMTWTP instance consists of $n=100$ jobs and is generated as explained in the following:
		For each job $j\in\{1,...,100\}$ a processing time $p_{j}$ is chosen uniformly at random from $\{1,...,100\}$ and a job
		weight $w_{j}$ is chosen uniformly at random from $\{1,...,10\}$. In addition, the due date $d_j$ of job $j$ is chosen uniformly at random from 
		\begin{equation}
			\nonumber
			\left[ \sum_{j = 1}^{n} p_{j} \cdot (1 - TF - \frac{RDD}{2}),
			\sum_{j = 1}^{n} p_{j} \cdot (1 - TF + \frac{RDD}{2}) \right],
			\label{eq:tf_rdd_func}
		\end{equation}
		where $TF$ and $RDD$ are parameters each from the set \linebreak$\left\{ 0.2, 0.4, 0.6, 0.8, 1.0 \right\}$.
		Parameter $TF$ represents the hardness of an SMTWTP instance and parameter $RDD$ represents the relative range of due dates.
		A great $TF$ value results in a small (or even negative) lower bound on the due dates. 
		Since due dates cannot be negative, all negative due dates are set to $0$.
		Observe that such a job is always completed after its due date, i.e., it contributes a positive tardiness for all schedules.
		In contrast a small $TF$ value results in large due dates and thus, more jobs can be expected to be finished in time.
		The variance of the due dates is determined by the parameter $RDD$.
		In particular, a great $RDD$ value results in more diverse due dates, whereas a small $RDD$ value results in  more similar due dates.
		
		For each combination of parameters $TF$ and $RDD$ five problem instances were generated.
		Consequently, a set of $125$ SMTWTP instances was generated.
		The set of all $125$ generated SMTWTP instances is called \emph{evaluation set} and it is henceforth denoted by $\mathcal{X}$.
		The subset of $\mathcal{X}$ that were generated with the parameter values $RDD=a$ and $TF=b$ is denoted by $\mathcal{X}_{a,b}$.	
		The evaluation set $\mathcal{X}$ was used for investigating the optimization behavior of the weighted population update rule.
		For that reason, we compared the results of the PACO algorithm with weighted population update rule (henceforth WPACO) with the results of the standard PACO. 
		In addition, the best-known solutions of all problem instances are used for the comparison.
		A listing of these solutions can be found in \cite{ORLIB}.
		It is worth mentioning that these solutions were obtained using the method that has been proposed in \cite{CPV2002}. 
		
		As the parameters $k$, $q_0$, $\alpha$, $\beta$, and $\tau_{max}$ have a crucial impact on the optimization behavior of ACO algorithms,
		a parameter optimization was conducted for the PACO and the WPACO in a two-stage procedure. First, sets of standard parameters were obtained
		from the literature. Then, for each algorithm an average initial best parameter setting was obtained by checking all combinations of the chosen
		parameters. In the second step, the automatic configuration tool \texttt{Irace} \cite{IRACE} was used to optimize the values of parameters $\alpha$ and $\beta$. Both steps are explained in detail in the following paragraphs.
		
		The parameters $q_{0} \in \left\{ 0.1, 0.5, 0.9 \right\}$, $k \in \left\{ 1, 5, 25 \right\}$, and 
		$\tau_{max} \in \left\{1, 3, 10 \right\}$ were obtained from the literature on PACO \cite{GM2002}.
		Recall that PACO and WPACO have different notions of the term population. 
		Therefore, we use the parameter $k_{\text{PACO}}\in \left\{ 1, 5, 25 \right\}$ and $k_{\text{WPACO}}\in \left\{ 10, 50, 100 \right\}$ to
		denote the size of the population in the respective algorithm. 
		The larger values of parameter $k_{\text{WPACO}}$ were chosen with respect to the maximum weight~$10$. 
		The reasoning is that all jobs within a multiset of the weighted population would be equivalent if the weight of a job is larger than the capacity of the weighted population. 
		In the first parameter optimization step, all problem instances were solved by the PACO and the WPACO algorithm using all $27$ combinations of the values of $q_0$, $\tau_{max}$, $k_{\text{PACO}}$, and $q_0$, $\tau_{max}$, $k_{\text{WPACO}}$, respectively. 
		Each problem instance from $\mathcal{X}$ was solved $5$ times for each value combination and each algorithm.
		The remaining parameter values of both algorithms were chosen as follows:
		The number of ants was $10$ and the number of iterations was 10000.
		In addition, standard values $\alpha=1$ and $\beta=2$ were used.
		The heuristic information was obtained using the modified MDD heuristic (Formula~\eqref{eq:mod_mdd}) and the solutions were constructed using the summation rule (formulas~\eqref{eq:nextjob_sum_max} and \eqref{eq:nextjob_sum_prob}). 
		These decisions are based on results presented in \cite{MM2000}.
		
		The aim of the second step of parameter optimization is to improve the initial parameter values that were obtained at the end of the first step.
		In particular, the values for parameters $\alpha$ and $\beta$ were optimized using the automatic configuration tool \texttt{Irace} \cite{IRACE}, which is an
		extension of the Iterated F-race procedure \cite{Birattari2010}. 
		Given a set of problem instances, an algorithm that solves these problems, and a set of parameters of the algorithm, the iterated racing procedure
		consists of three main phases that are iteratively performed until a stopping criterion is met: 
		First, new parameter configurations are selected from the parameter space according to a particular sampling distribution. 
		The initial parameter space is spanned by the ranges of the input parameters.
		Second, the best of these parameter configurations are determined according to a statistical approach.
		Third, the sampling distribution is adjusted in order to sample towards the best configurations.
		After the stopping criterion is met, \texttt{Irace} returns a set of most appropriated parameter settings for the given set of problem instances.
		For more information on iterated racing and the \texttt{Irace} tool, the reader is referred to \cite{IRACE}.
		\texttt{Irace} was used to optimize the values of parameters $\alpha, \beta \in [0.5,3.0]$ (with step size 0.001) for each problem instance of the evaluation set individually.
		Initial tests showed that the lower and upper bounds on $\alpha$ and $\beta$ are appropriate.
		\texttt{Irace} was configured to run each instance $2000$ times. 
		The step of parameter optimization is performed for the PACO algorithm only in order to achieve a clear competitive advantage for the PACO algorithm. 
		The idea is to show that the weighted population update rule is able to improve the solutions of the standard PACO algorithm even if the values of the parameters are not explicitly tuned for this algorithm.
		The outcome of the second parameter optimization step is that for each problem instance a most appropriated setting of parameter values is determined for the standard PACO algorithm.
		
		To evaluate the proposed population update rule a third experiment was conducted.
		Each problem instance of $\mathcal{X}$ was solved $5$ times by the PACO and the WPACO algorithm.
		For each computation, the parameter settings that were obtained by the use of \texttt{Irace} were used for both algorithms.
		
		Combining ACO algorithms with local search strategies has a high impact on solution quality.
		On one hand, it has been proven to improve the solution quality significantly, e.g., see \cite{BSD2000,MM2000} and \cite{Oliveira2011} for results on ACO  and PACO, respectively.
		On the other hand, it moves much of the optimization process away from the ACO algorithm.
		As the main objective of this work is to investigate the proposed population update rule, all presented PACO algorithms do not utilize local search strategies.
		
%
%
	\section{Results}
	\label{sec:results}
	
		\begin{figure}
			\centering
			\subfloat[][]{
				\includegraphics[width=0.25\textwidth]{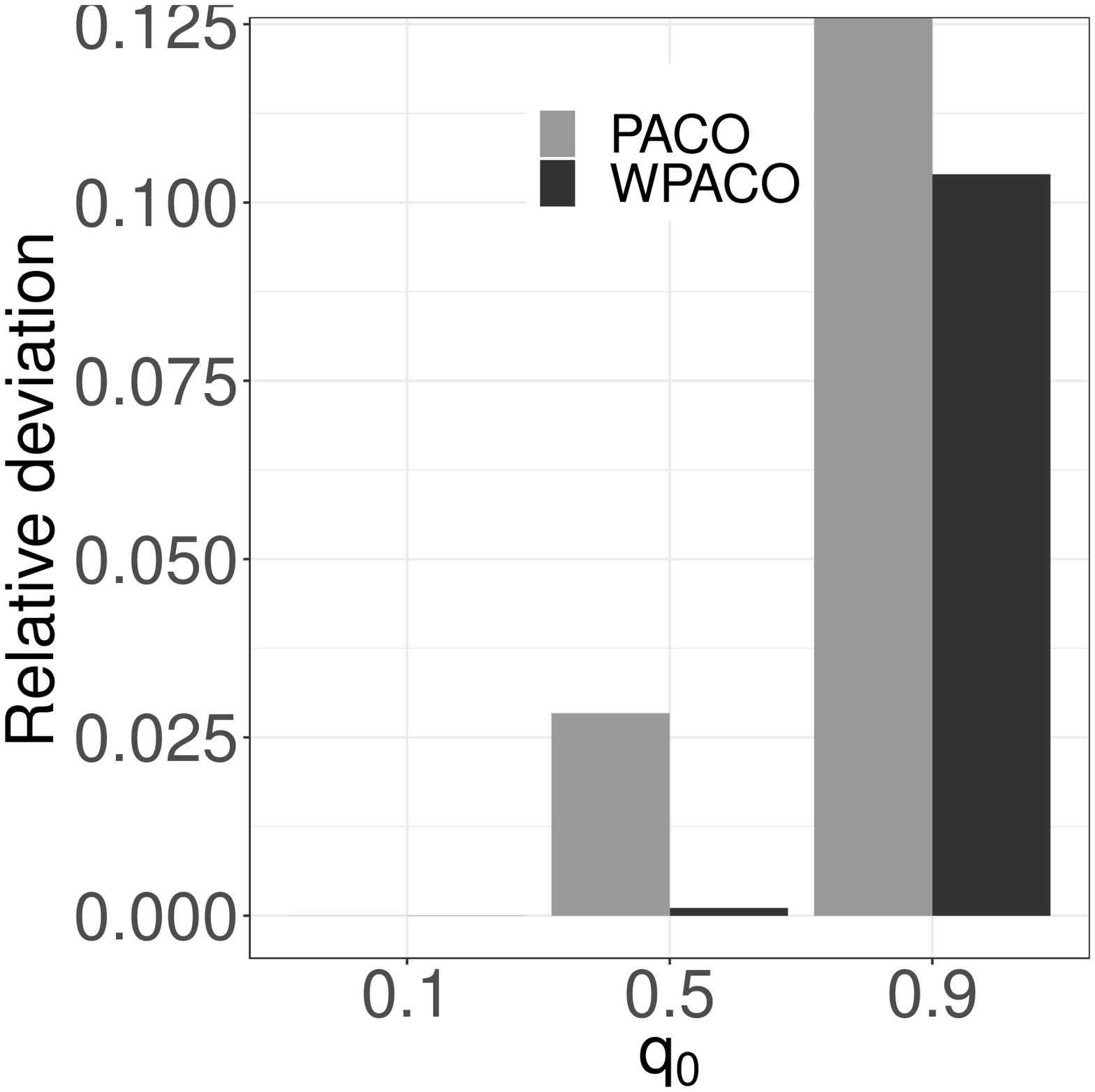}
			}
			\qquad
			\subfloat[][]{
				\includegraphics[width=0.25\textwidth]{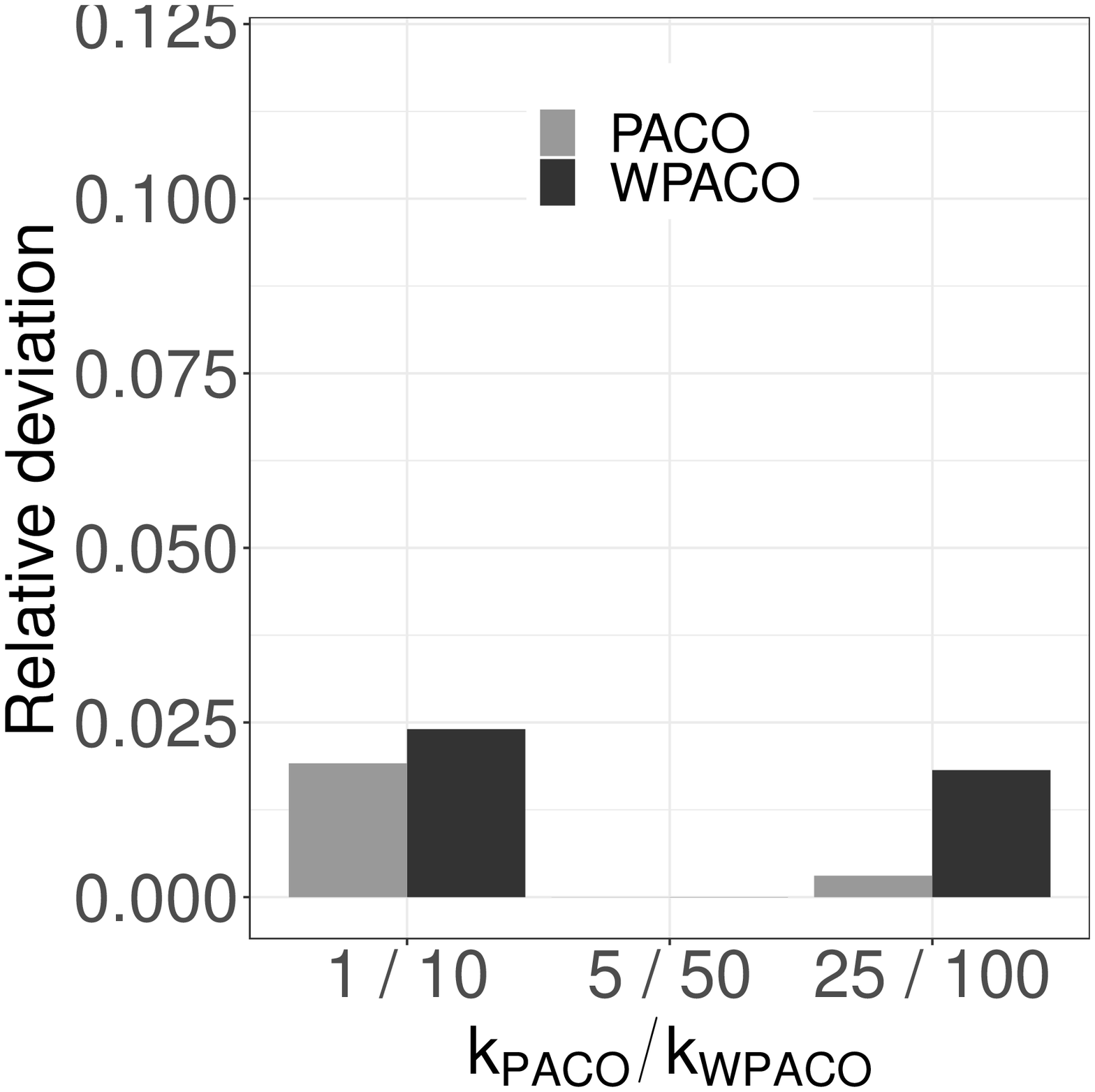}
			}
			\qquad
			\subfloat[][]{
				\includegraphics[width=0.25\textwidth]{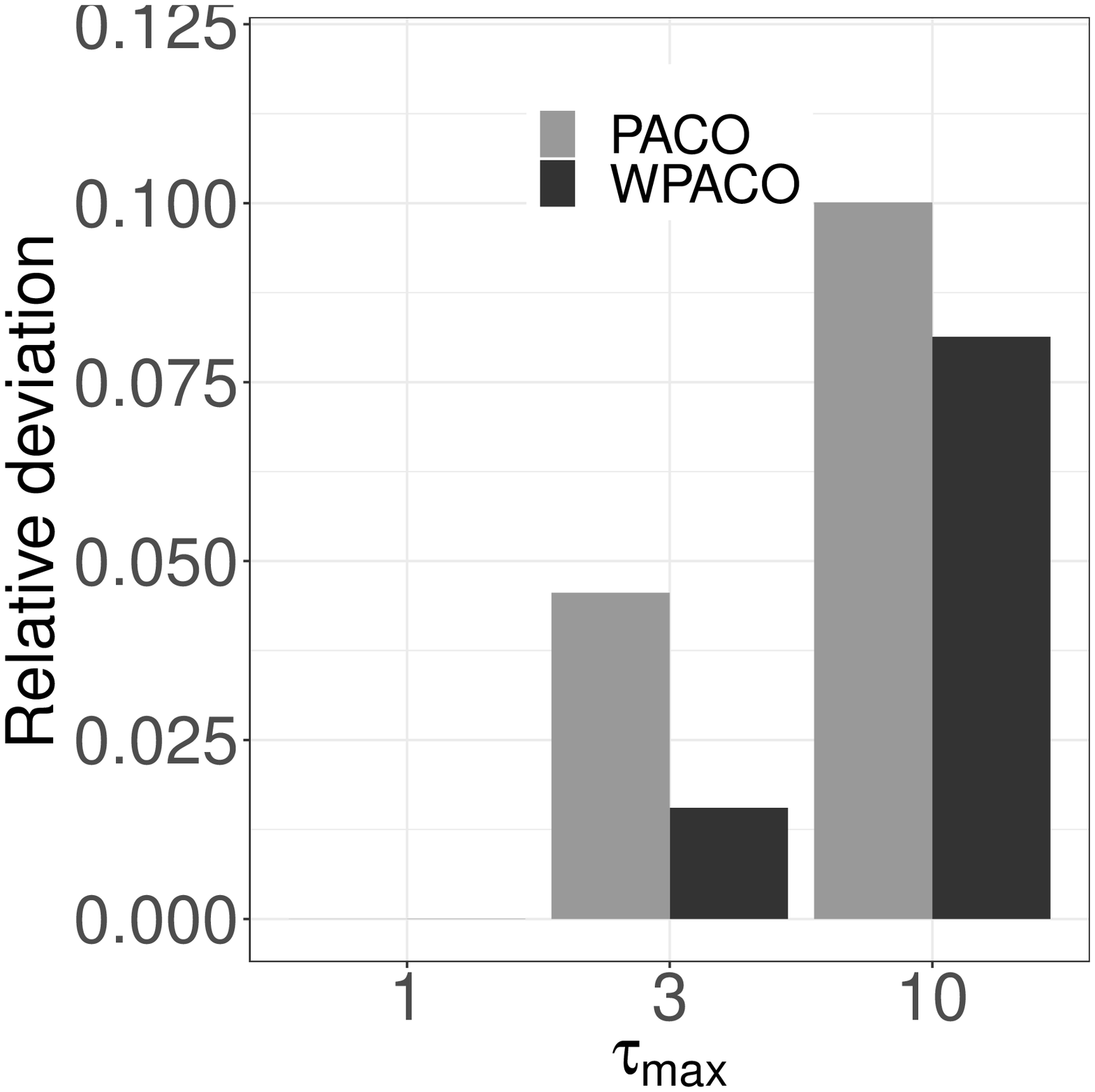}
			}
			\caption{
				Relative deviation from the average total weighted tardiness (TWT) achieved by applying the algorithms PACO and WPACO to the evaluation set $\mathcal{X}$.
				The results are illustrated for different parameter values of $q_{0}$ (a), $k_{\text{PACO}}/k_{\text{WPACO}}$ (b), and $\tau_{max}$ (c) and in relation to the parameter configuration that resulted in the smallest TWT, i.e., $q_{0}=0.1$, $\tau_{max}=1$, $k_{\text{PACO}}=5$, and $k_{\text{WPACO}}=50$.
				}
			\label{fig:diff_param_qkt}
		\end{figure}
		
		Figure~\ref{fig:diff_param_qkt} shows the average total weighted tardiness (TWT) achieved by the PACO and the WPACO algorithm for the first step of the parameter optimization.
		It can be seen that on average the best TWT is obtained for parameter values 
		$q_{0} = 0.1$, $\tau_{max} = 1$, $k_{\text{PACO}} = 5$, and $k_{\text{WPACO}} = 50$. 
		The results show that both algorithms, i.e., PACO and WPACO, achieve on average a smaller TWT for smaller values of $q_0$ and $\tau_{max}$.
		Whereas small values of $q_0$ enhance the exploration of different solutions, small values of parameter $\tau_{max}$ increase the influence of the heuristic information during the optimization process. 
		The reason is that a small $\tau_{max}$ value results in small pheromone values $\tau_{ij}$. 
		As the values $\eta_{ij}$ are independent of $\tau_{max}$, it holds by formulas~\eqref{eq:nextjob_sum_max} and \eqref{eq:nextjob_sum_prob} that the influence of the heuristic information increases for smaller pheromone values.
		The results for the population size parameters $k_{\text{PACO}}$ and $k_{\text{WPACO}}$ show that the best TWT is obtained by both algorithms with a medium-sized population. 
		This result agrees with results from the literature on PACO, e.g., see \cite{GM2002}.
		
		\begin{figure}
			\centering
			\subfloat[][]{
				\includegraphics[width=0.3\linewidth]{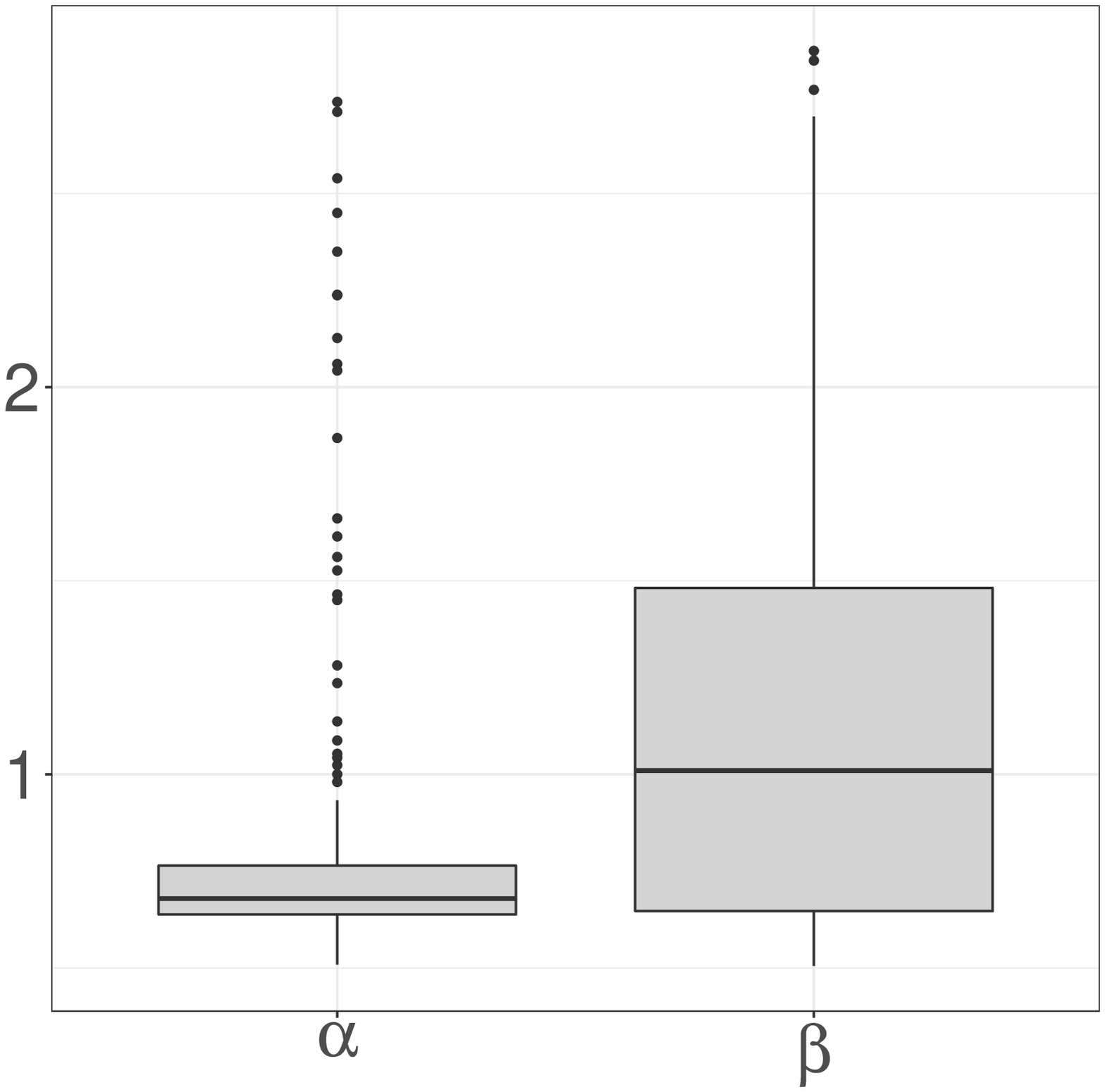}
			}
			\qquad
			\subfloat[][]{
				\includegraphics[width=0.3\linewidth]{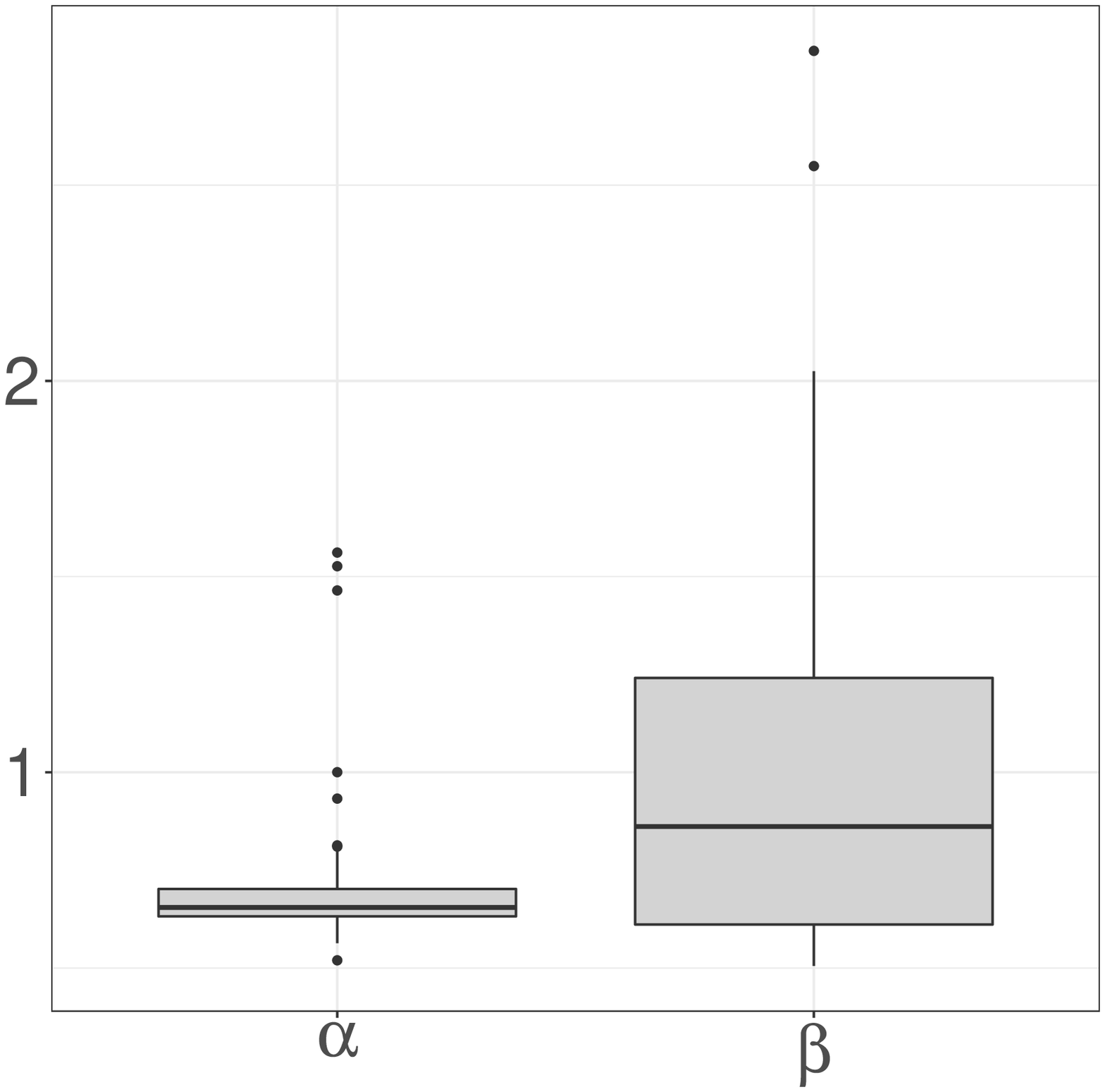}
			}				
			\caption{
				Distribution of parameters $\alpha$ and $\beta$ obtained by \texttt{Irace}.
				The boxplots show the optimized values of $\alpha$ and $\beta$ for all problem instances of the evaluation set $\mathcal{X}$ (a) and $\mathcal{X}\setminus (\mathcal{X}_{0.2,0.4}\cup\mathcal{X}_{0.2,0.6}\cup\mathcal{X}_{0.2,0.8}\cup \mathcal{X}_{0.2,1.0})$ (b).
			}
			\label{fig:alpha_beta_box}
		\end{figure}

		The aim of the second step of parameter optimization is to utilize the software tool \texttt{Irace} to further improve the parameter values used for the PACO algorithm.
		Parameter values $q_{0} = 0.1$, $k_{\text{PACO}} = 5$, and $\tau_{max} = 1$ were fixed during this step of parameter optimization. 
		The reason is that initial tests showed that optimizing these parameters on the evaluation set results in highly similar parameter configurations.
		As a consequence, the software tool \texttt{Irace} was used to optimize the values of parameters $\alpha$ and $\beta$ within the range $[0.5,3.0]$ only.
		
		Since \texttt{Irace} has been used to optimize the parameter values for each problem instance of the evaluation set separately, a distribution of optimized values for $\alpha$ and $\beta$ parameters has been obtained.
		Figure~\ref{fig:alpha_beta_box}.(a) illustrates this distribution.
		It can be seen that most optimized values for $\alpha$ and $\beta$ lay in the intervals  $\left[ 0.5, 0.76 \right]$ and  $\left[ 0.5, 1.48 \right]$, respectively.
		Figure~\ref{fig:alpha_beta_box}.(a) also shows a large number of outliers that mostly pertain to parameter $\alpha$. 
		Most of the outliers correspond to problem instances from $\mathcal{X}$ for which it holds that $RDD=0.2$ and  $TF \in \left\{ 0.4, 0.6, 0.8, 1.0 \right\}$, i.e., for the sets $\mathcal{X}_{0.2,0.4}$, $\mathcal{X}_{0.2,0.6}$, $\mathcal{X}_{0.2,0.8}$, and $\mathcal{X}_{0.2,1.0}$.
		The reason is that these problem instances could be solved optimally with nearly each combination of values for $\alpha$ and $\beta$.
		Consequently, problem instances from $\mathcal{X}_{0.2,0.4}\cup\mathcal{X}_{0.2,0.6}\cup\mathcal{X}_{0.2,0.8}\cup \mathcal{X}_{0.2,1.0}$ do not allow a parameter optimization.
		As a result, the $\alpha$ and $\beta$ values that were obtained from \texttt{Irace} by tuning the parameter values for PACO on these problem instances were removed from the distribution. The resulting distributions for $\alpha$ and $\beta$ are shown in Figure~\ref{fig:alpha_beta_box}.(b). The figure shows that now only a few outliers occur.
		
		\begin{figure}
			\centering
			\includegraphics[width=0.6\linewidth]{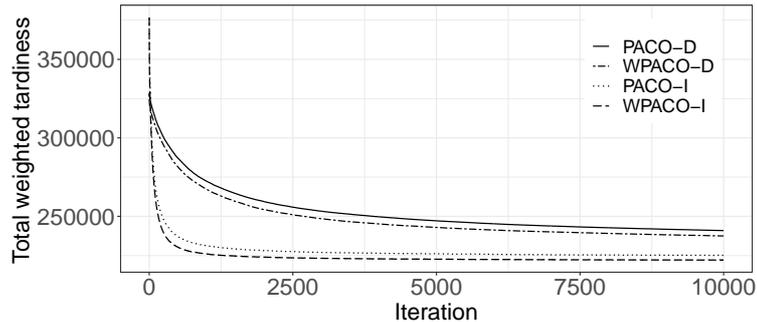}
			\caption{Average weighted tardiness over 10000 iterations computed for each problem instance of the evaluation set including 5 repetitions for each problem instance.
			PACO-D and WPACO-D use the default values $\alpha = 1$, $\beta = 2$ and the standard population update rule. Algorithms PACO-I and WPACO-I use the parameter configurations that were optimized by \texttt{Irace} and the proposed population update rule.}
			\label{fig:total_tardiness_irace_param}
		\end{figure}

		Over the course of 10000 iterations, Figure~\ref{fig:total_tardiness_irace_param} shows the average TWT achieved by applying the algorithms PACO and WPACO with and without optimized parameter values to each problem instance of the evaluation set. PACO and WPACO using the default values $\alpha = 1$, $\beta = 2$ are denoted by PACO-D and WPACO-D, respectively, and PACO-I and WPACO-I, respectively, for the values $\alpha$, $\beta$ that	were optimized by \texttt{Irace}.
		It can be seen that algorithms PACO-I and WPACO-I achieve solutions with significantly smaller average TWT than algorithms PACO-D and WPACO-D.
		Whereas algorithms PACO-I and WPACO-I converge approximately at iteration 1500, the algorithms PACO-D and WPACO-D converge considerably later around iteration 8000.
		The results show the benefit of the parameter optimization that was performed by \texttt{Irace}.
		Figure~\ref{fig:total_tardiness_irace_param} also shows the effect of the proposed population update rule: For each iteration, algorithm WPACO-I (WPACO-D) achieves solutions with smaller average TWT than PACO-I (respectively PACO-D).

		\begin{table}
			\centering
			\caption{Average total weighted tardiness of the best solutions from the OR-Library as well as the solutions obtained by applying PACO-D, PACO-I, WPACO-D, and WPACO-I to the problem instances from the evaluation set.}
			\label{tab:total_tardiness_vergleich}
			\begin{tabular}{|c|c|c|c|c|c|}
				\hline
				\textbf{Method} & \textbf{Total weighted tardiness} & \textbf{diff. to OR} \\\hline
				OR-Library & 217851 & - \\\hline
				PACO-D & 274140 & 25.8\% \\\hline
				WPACO-D & 268571 & 23.3\% \\\hline
				PACO-I & 233718 & 7.3\% \\\hline
				WPACO-I & 225641 & 3.6\% \\\hline
			\end{tabular}
		\end{table}

		The average TWT of the best solutions from the OR-Library as well as the solutions obtained by applying PACO-D, PACO-I, WPACO-D, and WPACO-I to the problem instances of the evaluation set are listed in Table~\ref{tab:total_tardiness_vergleich}.
		It shows that all four algorithms produce solutions that are on average worse than the best solutions of the OR-Library.
		This result is not surprising as all four algorithms are metaheuristics. 
		Another fact that contributes to the deviation from the average TWT of the best OR-Library solutions is that no local search strategy was used in order to improve already found solutions.
		However, Table~\ref{tab:total_tardiness_vergleich} also shows that the PACO algorithms that use the proposed weighted population update rule give much better results than PACO with the standard update rule.
		More precisely, algorithms WPACO-I and WPACO-D achieve solutions that exhibit an average TWT that is larger than the average TWT of the best OR-Library solutions by 3.6\% and 23.3\%, respectively.
		The corresponding PACO algorithms that use the standard update rule, i.e., PACO-I and PACO-D, obtain solutions with a TWT that is larger than the average TWT of the best OR-Library solutions by 7.3\% and 25.8\%, respectively.

		\begin{figure}
			\centering
			\includegraphics[width=0.9\textwidth]{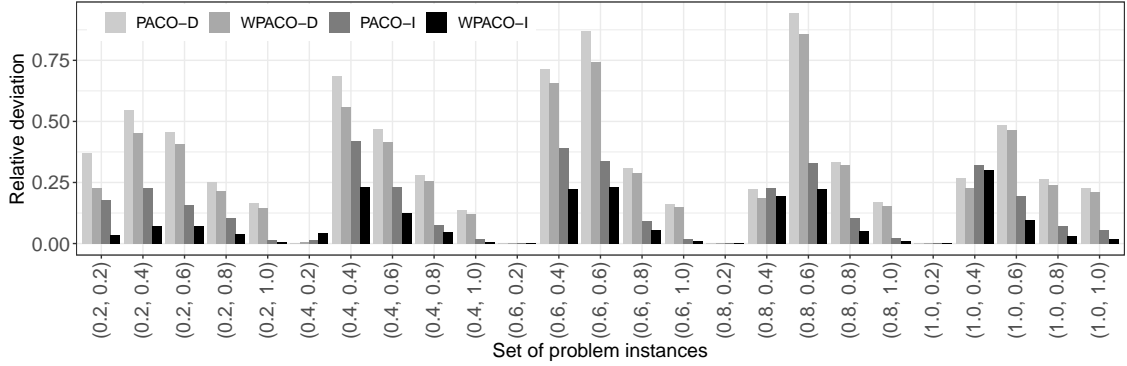}
			\caption{Relative deviation of the average weighted tardiness (y-axis) from the average weighted tardiness of the best OR-Library solutions for PACO-D, PACO-I, WPACO-D, and WPACO-I for the problem instances of the evaluation set. The relative deviation is illustrated for each algorithm and the solutions of each set $\mathcal{X}_{a,b}$ that was constructed using the parameters  $RDD=a$ and $TF=b$. The set $\mathcal{X}_{a,b}$ is represented by the notation $(a,b)$ (x-axis).}
			\label{fig:rdd_tf_ranges}
		\end{figure}

		Figure~\ref{fig:rdd_tf_ranges} shows by which fraction the average TWT obtained by all four investigated PACO algorithms deviates from the average TWT of the best OR-Library solutions with respect to all combinations of parameter values $TF$ and $RDD$ that were used for the construction of the evaluation set.
		Generally, it can be seen that the algorithms PACO-I and WPACO-I achieve much smaller relative deviations, i.e., a much smaller average TWT, than algorithms PACO-D and WPACO-D, respectively.
		The figure also shows that the best OR-Library solutions were found by each algorithm for the problem instances of the sets $\mathcal{X}_{0.2,0.6}$,  $\mathcal{X}_{0.2,0.8}$, and  $\mathcal{X}_{0.2,1.0}$. This result can be explained by the fact is that these combinations of $RDD$ and $TF$ values lead to comparatively large and diverse due dates.
		The problem instances of the evaluation set that were generated with $TF=0.2$ appear to become harder for smaller $RDD$ values.
		In particular, the problem instances from set $\mathcal{X}_{0.2,0.4}$ were solved optimally by the PACO-D algorithm only.
		Moreover, the problem instances from set $\mathcal{X}_{0.2,0.2}$ were not solved optimally by all investigated algorithms.
		One reason is that the due dates become less diverse for smaller values of $RDD$.	
		The worst average TWT was achieved for the problem instances from the sets $\mathcal{X}_{0.6,0.6}$ and $\mathcal{X}_{0.8,0.6}$.
		This result is consistent with the observations that were made in \cite{BSD2000,geiger2010heuristic} and which were explained and investigated in more detail in \cite{merkle2001new}.
		In particular, the authors stated that SMTWTP instances that were generated with $TF = 0.6$ appear to be difficult to solve.
		Figure~\ref{fig:rdd_tf_ranges} also shows how advantageous it is for the PACO to use the proposed weighted population update rule.
		For all problem instances from $\mathcal{X}\setminus\mathcal{X}_{0.2,0.4}$ the algorithms WPACO-I and WPACO-D achieve a much smaller average TWT than the algorithms PACO-I and PACO-D, respectively.
	
		\begin{figure}
			\centering
			\includegraphics[width=0.7\linewidth]{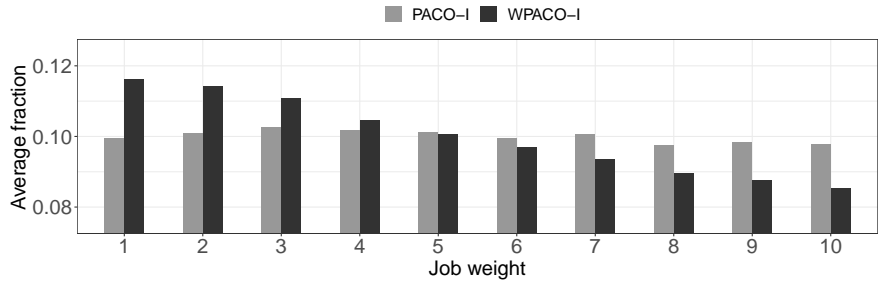}
			\caption{
				Average fraction of iterations where a job with a certain weight changes its position within the iteration best schedules.}
			\label{fig:verteilung_jobwichtung_aenderung_paco_wpaco}
		\end{figure}
		
		\begin{figure}
			\centering
			\includegraphics[width=0.6\linewidth]{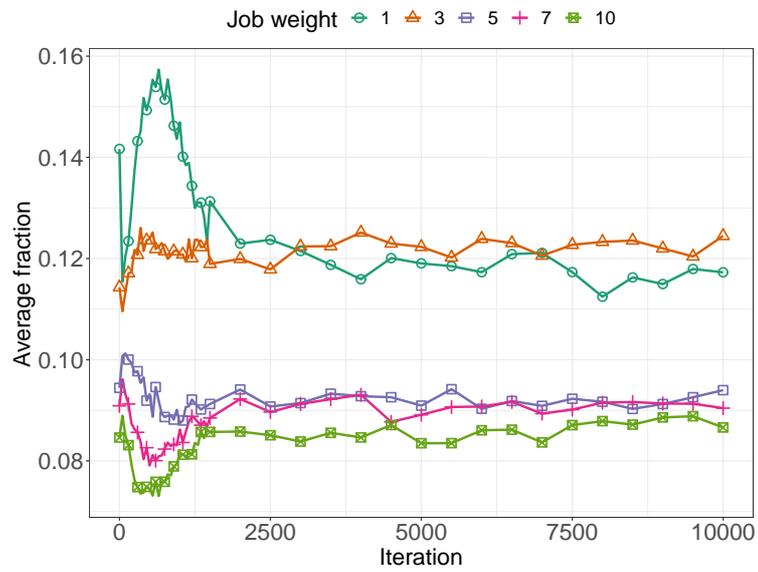}
			\caption{Average fraction that a job with a certain weight changes its position within successive iteration best schedules over 10000 iterations by applying 
			WPACO-I. For better visualization only job weights 1, 3, 5, 7 and 10 are included.}
			\label{fig:avg_fraction_iteration}
		\end{figure}
		
		\begin{figure}
			\centering
			\includegraphics[width=0.6\linewidth]{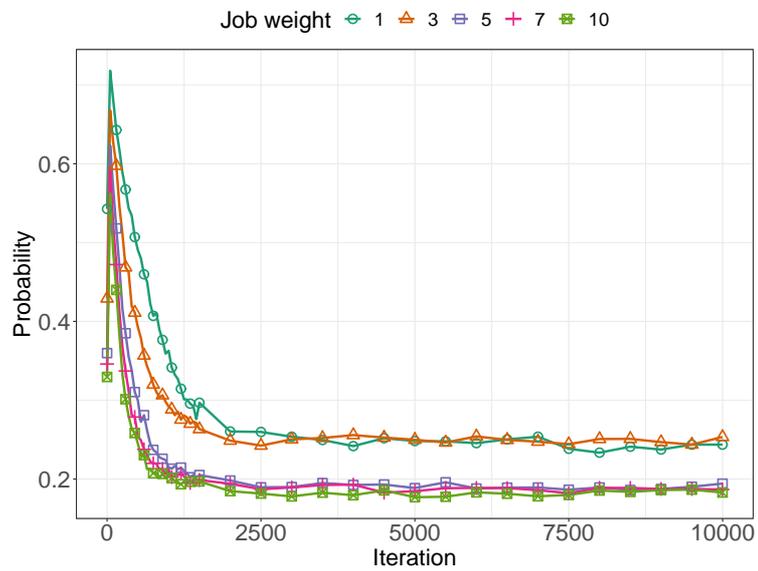}
			\caption{Empirical probability that a job with a certain weight changes its position within the iteration best schedules over 10000 iterations by applying WPACO-I. For better visualization only job weights 1, 3, 5, 7 and 10 are included.}
			\label{fig:job_change_prob_iteration}
		\end{figure}

		To investigate the difference between the weighted population update rule and the standard population update rule with respect to the composition of a population during optimization, Figure~\ref{fig:verteilung_jobwichtung_aenderung_paco_wpaco} shows how often a job with a certain weight changes its position in a single iteration.
		For each problem instance of the evaluation set and the algorithms PACO-I and WPACO-I, the figure was obtained by comparing the iteration best schedules of successive iterations.
		For PACO-I it can be seen that the average fraction of iterations is equally distributed among the job weights. The reason for this is that PACO-I uses the standard population update rule which cannot consider the weights of the jobs a given SMTWTP instance.
		For WPACO-I the figure shows that the average fraction of a position change increases for a decreasing job weight. Thus, jobs with a large weight are less likely to get scheduled to another position. The reason for this is explained in the following. Jobs with a small weight have less influence on the TWT than jobs with a large weight. Hence, jobs with a large weight are scheduled early during optimization in order to reduce the TWT of a schedule.
		Since the weighed population update rule adds jobs with a large weight multiple times to the population, it follows that future ants prefer the same position for those jobs.
		A consequence is that the position of a job with a large weight is fixed in early iterations which reduces its average fraction of position change illustrated in Figure~\ref{fig:verteilung_jobwichtung_aenderung_paco_wpaco}.
		In the following iterations, the process of optimization focuses on jobs with smaller weights leading to an increased average fraction of their position changes.
		Figure~\ref{fig:avg_fraction_iteration} displays this result. After a short initialization phase the average fraction of jobs with smaller weights increases significantly.
		After approximately 1500 iterations, the average fractions adjust. At this point the
		algorithm converges, as Figure~\ref{fig:total_tardiness_irace_param} shows.
		Additionally, the probability that a job changes its position decreases in further iterations. Figure~\ref{fig:job_change_prob_iteration} illustrates this effect. After the algorithm
		converges (approximately 1500 iterations), the probabilities become stable.
		Altogether this shows that for WPACO-I, which uses the weighted population update rule, a correlations becomes noticeable that jobs with large weights are less likely to get rescheduled at another position. 
		To verify the assumption, the Pearson correlation between the job weight and the average fraction of iterations were its position changes was calculated.
		The result is a probability $p = 6.286 \cdot 10^{-9}$ and a correlation coefficient $r = -0.994$. This verifies the strong negative correlation.

%
%
	\section{Conclusions}
	\label{sec:conc}
	In this paper a novel population update rule for population based ant colony optimization (PACO) has been presented for the example of the single machine total weighted tardiness problem.
	The new update rule, called weighted population update rule, allows to weight different parts of a solution.
	PACO with the new population update rule (WPACO) has achieved better solution quality for 125 benchmark problem instances  than its counterpart that used the standard population update rule.
	A detailed analysis of the solutions obtained by WPACO has revealed a strong negative correlation between the weight of a job and the probability that a job gets rescheduled at another position in successive iterations.
    
    For future work it is planned to explore the possibility to apply WPACO to other optimization problems like  traveling salesperson problems or quadratic assignment problems.
%
%
\bibliographystyle{plain}
\bibliography{WPACOPaper}

\end{document}